\documentclass{article}
\pdfoutput=1
\usepackage{PRIMEarxiv}

\usepackage[utf8]{inputenc} 
\usepackage[T1]{fontenc}    
\usepackage{hyperref}       
\usepackage{url}            
\usepackage{booktabs}       
\usepackage{amsfonts}       
\usepackage{nicefrac}       
\usepackage{microtype}      
\usepackage{lipsum}
\usepackage{float}
\usepackage{fancyhdr}       
\usepackage{graphicx}       
\usepackage{authblk}        
\usepackage{amsmath}        
\usepackage{lineno}
\usepackage{multirow}
\usepackage{svg}

\graphicspath{{media/}}     
\usepackage{parskip}
\newcommand{\beginsupplement}{%
        \setcounter{table}{0}
        \renewcommand{\thetable}{S\arabic{table}}%
        \setcounter{figure}{0}
        \renewcommand{\thefigure}{S\arabic{figure}}%
     }

\title{Evaluating the Role of Training Data Origin for Country-Scale Cropland Mapping in Data-Scarce Regions: A Case Study of Nigeria}

\author[1,4,*]{\textbf{Joaquin Gajardo}}
\author[2]{\textbf{Michele Volpi}}
\author[1]{\textbf{Daniel Onwude}}
\author[1,3]{\textbf{Thijs Defraeye}}
\affil[1]{Empa, Swiss Federal Laboratories for Material Science and Technology. Laboratory for Biomimetic Membranes and Textiles. Lerchenfeldstrasse 5, CH-9014 St. Gallen, Switzerland}
\affil[2]{SDSC, Swiss Data Science Center, ETH Zurich and EPFL, Switzerland}
\affil[3]{Food Quality and Design, Wageningen University \& Research, P.O. Box 17, 6700 AA Wageningen, the Netherlands}
\affil[4]{Institute of Agricultural Sciences, ETH Zurich, Universitätstrasse 2, Zurich, 8092, Switzerland}
\affil[*]{\emph{Corresponding author. Now at ETH Zurich. Email: jgajardo@ethz.ch}}

\begin{document}
\maketitle

\begin{abstract}

Cropland maps are essential for remote sensing-based agricultural monitoring, providing timely insights about agricultural development without requiring extensive field surveys. While machine learning enables large-scale mapping, it relies on geo-referenced ground-truth data, which is time-consuming to collect, motivating efforts to integrate global datasets for mapping in data-scarce regions. A key challenge is understanding how the quantity, quality, and proximity of the training data to the target region influences model performance in regions with limited local ground truth. To address this, we evaluate the impact of combining global and local datasets for cropland mapping in Nigeria at 10 m resolution. We manually labelled 1,827 data points evenly distributed across Nigeria and leveraged the crowd-sourced Geowiki dataset, evaluating three subsets of it: Nigeria, Nigeria + neighbouring countries, and worldwide. Using Google Earth Engine (GEE), we extracted multi-source time series data from Sentinel-1, Sentinel-2, ERA5 climate, and a digital elevation model (DEM) and compared Random Forest (RF) classifiers with Long Short-Term Memory (LSTM) networks, including a lightweight multi-task learning variant (multi-headed LSTM), previously applied to cropland mapping in other regions. Our findings highlight the importance of local training data, which consistently improved performance, with accuracy gains up to 0.246 (RF) and 0.178 (LSTM). Models trained on Nigeria-only or regional datasets outperformed those trained on global data, except for the multi-headed LSTM, which uniquely benefited from global samples when local data was unavailable. A sensitivity analysis revealed that Sentinel-1, climate, and topographic data were particularly important, as their removal reduced accuracy by up to 0.154 and F1-score by 0.593. Handling class imbalance was also critical, with weighted loss functions improving accuracy by up to 0.071 for the single-headed LSTM. Our best-performing model, a single-headed LSTM trained on Nigeria-only data, achieved an F1-score of 0.814 and accuracy of 0.842, performing competitively with the best global land cover product and showing strong recall performance, a metric highly-relevant for food security applications. These results underscore the value of regionally focused training data, proper class imbalance handling, and multi-modal feature integration for improving cropland mapping in data-scarce regions. We release our data, source code, output maps, and an interactive GEE web application to facilitate further research.

\end{abstract}


\begin{figure}[htbp]
  \centering
  \includegraphics[scale=0.35]{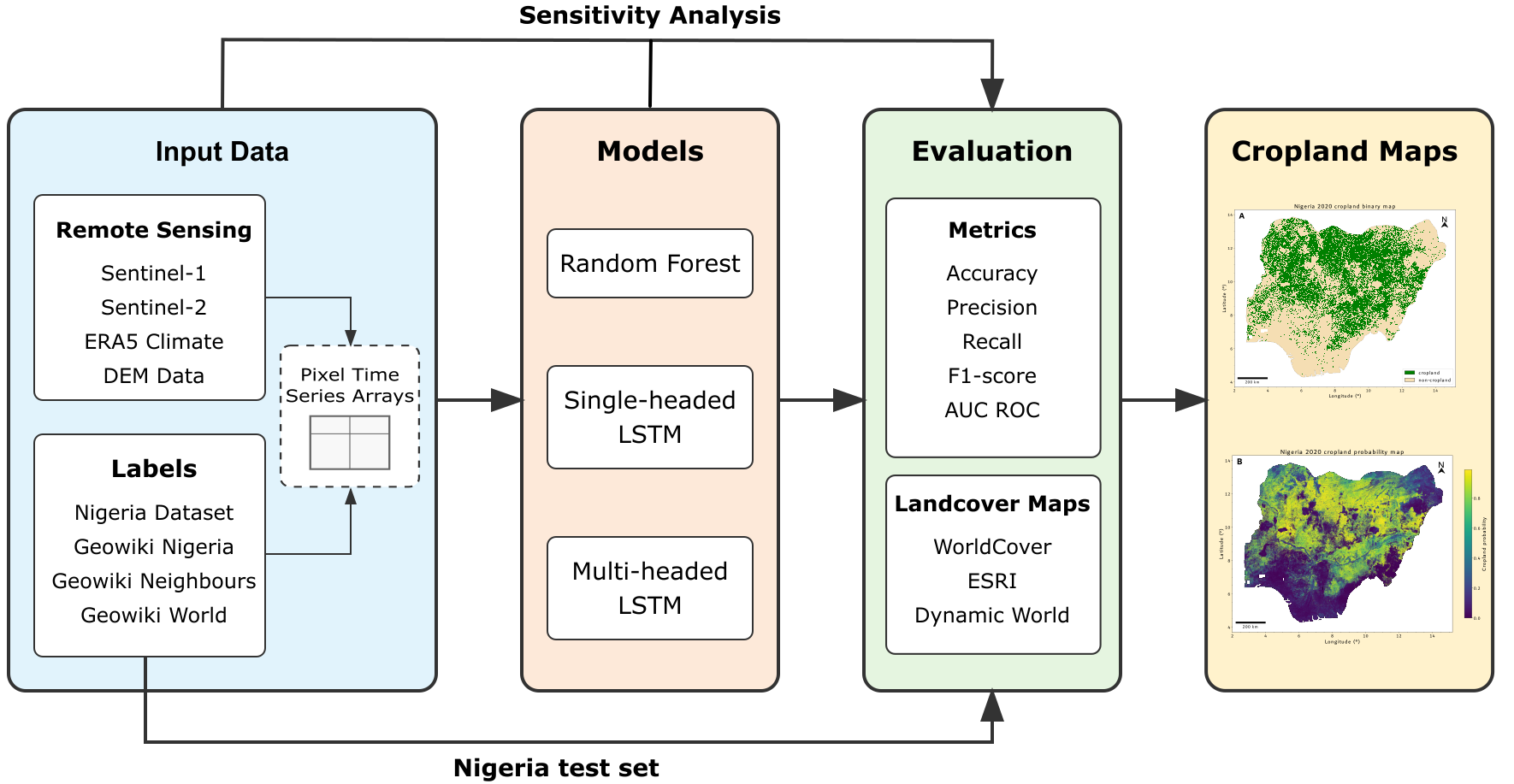}
  \caption{Schematic diagram of our methodology for cropland mapping in Nigeria. Input data combines remote sensing observations (Sentinel-1, Sentinel-2, ERA5 Climate data, and topographic data) with labels from various geographical extents—local (Nigeria-only), regional (Nigeria and neighbouring countries), and global (Geowiki World)—processed into monthly pixel time series arrays of a full year. Three different models are utilized: RF and two LSTM variants (single-headed and multi-headed). Model performance is evaluated using standard metrics and compared to existing land cover products, using a held-out Nigeria test set. The pipeline outputs binary and probability cropland maps for Nigeria.}
  \label{fig:diagram_pipeline}
\end{figure}

\section{Introduction}
\label{sec:intro}
Agricultural maps are relevant for a variety of downstream applications ranging from policy making, early-warning systems for food security, and agricultural extension services \cite{becker-reshef_crop_2023}. The availability of free, granular and high-cadence data from remote sensing sources like Sentinel-1 and 2 satellites has enabled the creation of detailed land cover and crop maps on a global scale \cite{venter_global_2022}. Machine learning (ML) is a widely employed methodology for generating these maps, given its capacity to process and extract useful information from large and diverse amounts of data, thereby aiding decision-making in agriculture \cite{wang_review_2022}.

Deep learning (DL) has significantly advanced crop mapping by overcoming the limitations of traditional ML approaches, which often rely on handcrafted features and struggle with temporal complexity \cite{wang_review_2022, PELLETIER2016156}. Convolutional Neural Networks (CNNs) were among the first DL models applied to crop classification, demonstrating superior performance over traditional Random Forest (RF) and Support Vector Machine (SVM) models \cite{7891032}. Recurrent Neural Networks (RNNs), particularly Long Short-Term Memory (LSTM) networks, were then introduced to capture crop phenology dynamics \cite{russwurm_temporal_2017}, later evolving into hybrid CNN-RNN architectures that leverage both spatial and temporal information \cite{turkoglu_crop_2021}. Other recent approaches include 3D CNNs and related 3D U-Nets, which process multi-temporal image cubes \cite{ji_3d_2018, adrian_sentinel_2021}, self-attention mechanisms for long-range temporal dependencies \cite{Garnot2019SatelliteIT, ruswurm_self-attention_2020}, and transformer-based models for improved spatio-temporal feature extraction from satellite time-series data \cite{li_multi-branch_2022}. Neural Differential Equations (NDEs) have also been explored to handle irregular sampling due to cloud cover in satellite imagery by implicitly modelling the continuous phenological development of crops \cite{metzger_crop_2021, gajardo2021neural}. Additionally, the integration of Sentinel-1 Synthetic Aperture Radar (SAR) data and Sentinel-2 optical imagery can significantly enhance classification accuracy and robustness to cloud cover with different ML and DL models \cite{adrian_sentinel_2021, garnot_multi-modal_2022}.  More recently, new augmentations have been explored, such as temporal random masking to hybrid LSTM-attention networks, which improved crop classification robustness to missing satellite observations \cite{zhang_improving_2024}. 

Using these models to generate crop maps in regions with limited ground-truth data remains a major challenge, as collecting sufficient high-quality training data requires significant time and resources. To mitigate this, researchers have explored leveraging geo-referenced cropland data from other regions, but severe data distribution shifts due to differences in climate, agricultural practices, and crop types complicate this approach. Transfer learning has emerged as a promising solution, using a variety of approaches such as multi-task learning \cite{kerner_rapid_2020, tseng_gabriel_annual_2020}, meta-learning \cite{tseng_learning_2021, tseng2022timl} and representation learning \cite{tseng_lightweight_2024}, aiming to improve model generalization across diverse agricultural landscapes. Despite these advances, applying these methods in new regions remains challenging, particularly where labelled datasets are scarce and local agricultural conditions differ from those in HICs (High-Income Countries), which concentrate most of the publicly available ground-truth data. A key open question is how globally distributed training data, considering its quantity, quality, and proximity to the target region, affects model performance in areas with limited local ground-truth. In this study, we address this question by focusing on cropland mapping in Nigeria, framing it as a crop versus non-crop binary classification problem. Nigeria is one of Africa’s leading food producers, accounting for nearly half of West Africa’s population \cite{ONWUDE202355}, however, it currently lacks a recent high-resolution cropland map. Thus, our goal was to produce a 10 m resolution cropland map for Nigeria using Sentinel data from the year 2020. To this end, we investigated two main research questions:

\begin{itemize}
\item What are the key technical components for effective cropland mapping in Nigeria, particularly in terms of model architecture, input features, and training strategies?
\item How does the geographical origin of training data (local, neighbouring countries, global) impact cropland mapping performance in Nigeria, and what is the best strategy to combine these sources?
\end{itemize}

To address these specific questions, we generated a small hand-labelled cropland dataset for Nigeria through expert photo-interpretation of high-resolution satellite and airborne imagery, and combined it with the global crowd-sourced Geowiki cropland dataset \cite{geowiki_laso_bayas_global_2017}. We compared an LSTM model to a RF baseline using multi-temporal inputs from Sentinel-1 and Sentinel-2 imagery, complemented by climate and topographic data (see Section ~\ref{sec:satellite_images}). For the LSTM, we explored both standard and multi-task learning approaches \cite{kerner_rapid_2020, tseng_gabriel_annual_2020}. Additionally, we tested different dataset configurations, including restricting the Geowiki dataset to Nigeria and neighbouring countries, analyzing the impact of input data modalities and class imbalance handling, and comparing our results to three existing global 10 m land cover maps from the same year (Section \ref{sec:landcover_maps}). A schematic summary of our study is presented in Figure ~\ref{fig:diagram_pipeline}.

\section{Data}
\subsection{Remote sensing data}
\label{sec:satellite_images}
Sentinel-1 and Sentinel-2 satellites from ESA's Copernicus program provide data at ground sample distance (GSD) ranging from 10 to 60 m with frequent revisit time of 5 days world-wide, which is the best resolution freely available. On the one hand, Sentinel-2 captures multispectral data in bands ranging from visible and near-infrared (GSD of 10 m) to short-wave infrared (20 m to 60 m GSD), which are useful for vegetation monitoring and crop detection, by measuring how plants reflect different wavelengths \cite{VUOLO2018122}. On the other hand, Sentinel-1 contains a SAR sensor operating in C-band (5.405 GHz), which measures actively emitted radio wave signals in two polarization modes, as reflected by the surface. SAR data is useful for detecting traits related to surface roughness such as crop height, growth stage and soil moisture content \cite{Sun2018_SAR}. In addition, we included pixel-wise data of monthly total precipitation, monthly average temperature at 2 metres above the ground from ERA5 Climate Reanalysis Meteorological data \cite{ERA5_data}, as well as elevation and slope derived from the Shuttle Radar Topography Mission’s (SRTM) Digital Elevation Model (DEM) \cite{SRTM_data}. 

We collected the data using the CropHarvest Python package (version 0.6.0) \cite{tseng2021cropharvest} and Google Earth Engine (GEE) \cite{GEE_GORELICK201718}, and produced a monthly per-pixel data array (i.e. a time series of 12 values for each pixel's features) for a requested year. We used all Sentinel-2 bands and up-sampled them to 10 m resolution using nearest neighbour interpolation (default in GEE), with exception to the 60 m resolution bands B1 and B10, which are predominantly meant for coastal aerosols and cirrus detection. We obtained least-cloudy monthly composites of Sentinel-2 Level-1C images using an algorithm from \cite{schmitt_aggregating_2019}. For Sentinel-1, we collected both VV and VH polarization bands from Ground Range Detected (GRD) products in Interferometric Wide (IW) swath mode, using images from a single orbit direction (ascending or descending) for each location to ensure consistency. Additionally, the we added the Normalized Difference Vegetation Index (NDVI = $\frac{NIR-R}{NIR+R}$), derived from Sentinel-2 red (B4) and near-infrared (B8) bands, given its value for assessing vegetation health and growth \cite{Sonobe2018_NDVI}. Thus, the output datasets contain modified Copernicus Sentinel data, ERA5, and SRTM data as processed by \cite{tseng2021cropharvest}, yielding samples that are 18-dimensional arrays of features at 12  different monthly time steps, counting backwards from the respective label collection date. We summarise the 18 considered features on Table \ref{tab:features}.

\begin{table}[htbp]
\small
\centering
\caption{Data features used in this study. We collected all features using the CropHarvest Python package \cite{tseng2021cropharvest}.}
\begin{tabular}{@{}llcll@{}}
\toprule
Feature group & Feature descriptions                      & Feature count  & Data type   & Source      \\ \midrule
SAR    & VV polarization mode                      & 1              & Dynamic     & Sentinel-1  \\
              & VH polarization mode                      & 1              & Dynamic     & Sentinel-1  \\ \midrule
Multispectral    & Visible bands (RGB)                       & 3              & Dynamic     & Sentinel-2  \\
              & Visible and Near Infrared bands (VNIR)    & 5              & Dynamic     & Sentinel-2  \\
              & Short Wave Infrared bands (SWIR)          & 3              & Dynamic     & Sentinel-2  \\
              & NDVI                                      & 1              & Dynamic     & Sentinel-2  \\ \midrule
Climate       & Monthly average precipitation             & 1              & Dynamic     & ERA5 \cite{ERA5_data}       \\
              & Monthly average temperature               & 1              & Dynamic     & ERA5 \cite{ERA5_data}       \\ \midrule
Topological   & Elevation                                 & 1              & Static      & SRTM \cite{SRTM_data}       \\
              & Slope                                     & 1              & Static      & SRTM \cite{SRTM_data}        \\ \bottomrule
\end{tabular}
\label{tab:features}
\end{table}

\subsection{Cropland labels}
\label{sec:data_labels}
Ground-truth data consists of manually annotated spatial coordinates that indicate the presence (or absence) of visible cropland at a specific time. Each data point includes latitude and longitude coordinates, a reference date, and a label indicating whether it represents cropland (1) or non-cropland (0). We used these coordinates and reference dates to retrieve corresponding remote sensing data.

\subsubsection{Geowiki dataset}
\label{sec:geowiki_datasets}

The Geowiki dataset consists of crowd-sourced points of cropland and non-cropland samples collected worldwide \cite{geowiki_laso_bayas_global_2017}. The dataset includes approximately 36,000 points, each associated with a cropland probability score derived from multiple human annotators. The data is readily available through the CropHarvest package \cite{tseng2021cropharvest}, with arrays covering September 2016 to September 2017, matching the labels' collection date to avoid mismatches due to possible land cover changes. We note that only 24,761 data arrays from the Geowiki labels are available within the CropHarvest package, reportedly due to missing satellite data at the different label locations (mainly Sentinel-1 time series) \cite{tseng2021cropharvest}. For this study, we used three configurations of the Geowiki data: the full global dataset (Geowiki World), samples from within Nigeria's boundaries (Geowiki Nigeria), and samples from Nigeria and its neighbouring countries (Geowiki Neighbours). We describe the detailed composition of these datasets and their processing for our experiments in Section~\ref{sec:dataset_preprocessing}.

\subsubsection{Nigeria dataset}
\begin{figure}[htbp]
  \centering
  \includegraphics[scale=0.3]{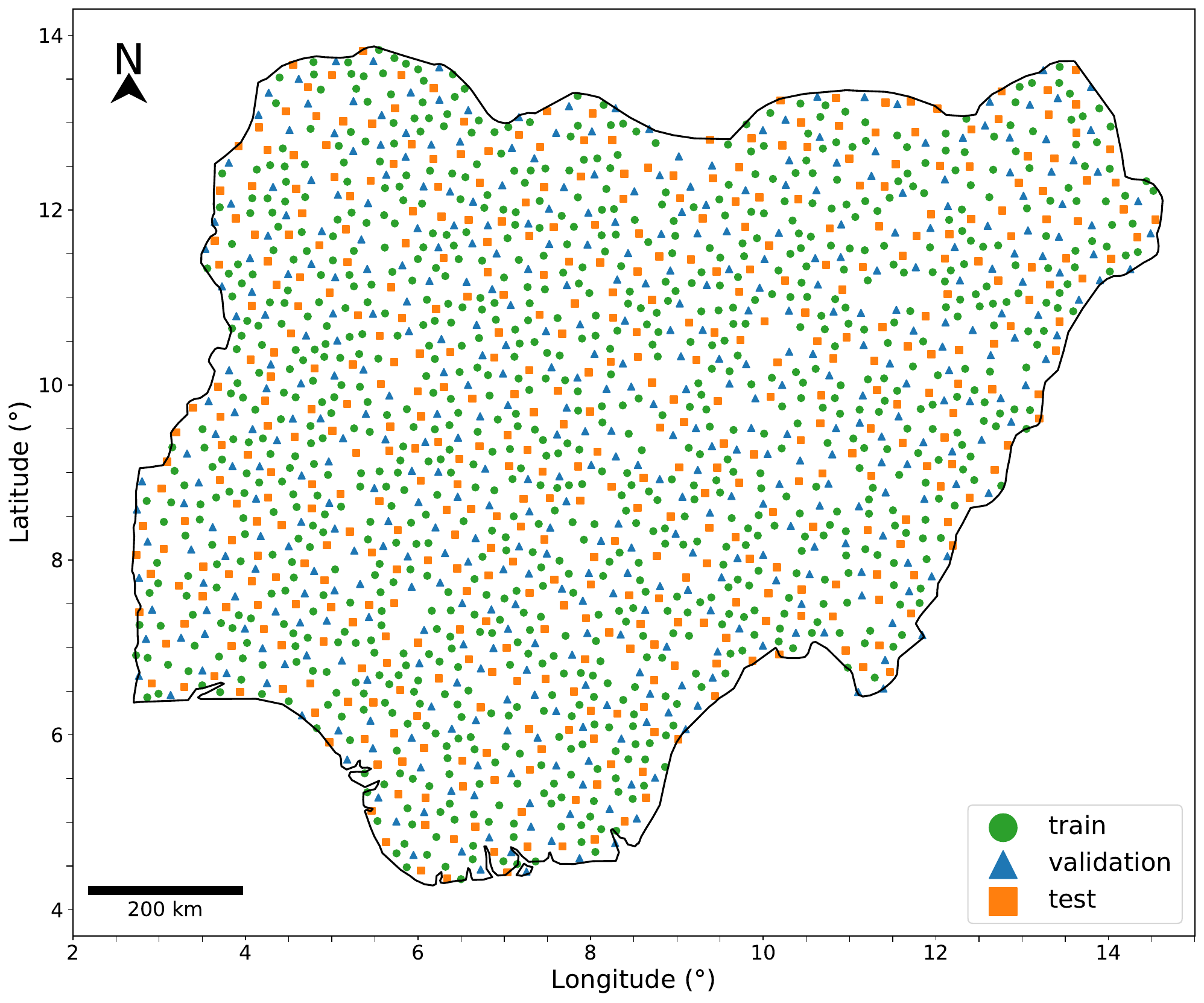}
  \caption{Geographical distribution of the 1,827 points of the generated Nigeria dataset and its assigned splits for model training and evaluation. Nigeria boundaries are taken from \cite{nigeria_boundaries}.}
  \label{fig:nigeria_data_split}
\end{figure}

We generated a dataset of evenly distributed points within Nigeria and labelled them as cropland or non-cropland through photo-interpretation. We randomly sampled spatial coordinates within the country until we obtained 2,000 points spaced by at least 15 km. An expert examined each point using very high-resolution Google Satellite Imagery Basemap (based on Maxar 30 cm GSD mosaics) from 2020 in Google Earth Pro and assigned them labels of cropland (1) or non-cropland (0) using QGIS \cite{QGIS_software}. We consider cropland as arable land (cultivated or fallow) and areas where permanent (perennial) crops are grown, including orchards \cite{FAO-2008_cropland_definition, perez-hoyos_comparison_2017}. This yielded 1,827 labelled points and the remaining were discarded as the class could not be confidently determined from the available reference imagery. We then used these point coordinates to query the respective pixel-wise data arrays from March 2019 - March 2020 using the CropHarvest package. Figure~\ref{fig:nigeria_data_split} shows the geographical distribution of the Nigeria dataset.

\subsubsection{Togo evaluation dataset}
\label{sec:togo_dataset}
We additionally used cropland labels from Togo, as provided in the CropHarvest evaluation dataset \cite{kerner_rapid_2020, tseng2021cropharvest}, to assess the performance of three global land cover maps (introduced in the following section \ref{sec:landcover_maps}) in comparison to published results from other country-specific cropland models. The test dataset from Togo comprises 306 labels covering the whole country, with 200 non-crop and 106 crop examples \cite{kerner_rapid_2020}.

\subsection{Reference land cover maps}
\label{sec:landcover_maps}
We used three global land cover maps from 2020 at 10 m resolution as reference maps for comparison: Google's Dynamic World \cite{brown_dynamic_2022}, ESRI's 2020 Land Cover \cite{ESRI_LULC_map_karra_global_2021} and ESA's WorldCover 2020 \cite{ESA_WorldCover_zanaga_daniele_2021_5571936}. Dynamic World and ESRI Land Cover were produced with DL models trained on a large dataset of 5 billion densely annotated patches of Sentinel-2 imagery distributed all over the world \cite{venter_global_2022, tait_dynamic_2021}. WorldCover was produced with a Catboost model \cite{dorogush_catboost_2018} trained on pixels from 100 m$^2$ patches of 141,000 locations around the world, using 131 features derived from Sentinel-1, Sentinel-2, and DEM data, as well as spatial features from biome data \cite{venter_global_2022, ESA_WorldCover_zanaga_daniele_2021_5571936}. WorldCover uses a smaller Minimum Mapping Unit (MMU) of 100 m$^2$ compared to 250 m$^2$ for Dynamic World and ESRI Land Cover.

\section{Methods}
\subsection{Data preprocessing}
\label{sec:dataset_preprocessing}

\begin{figure}[htbp]
  \centering
  \includegraphics[scale=0.4]{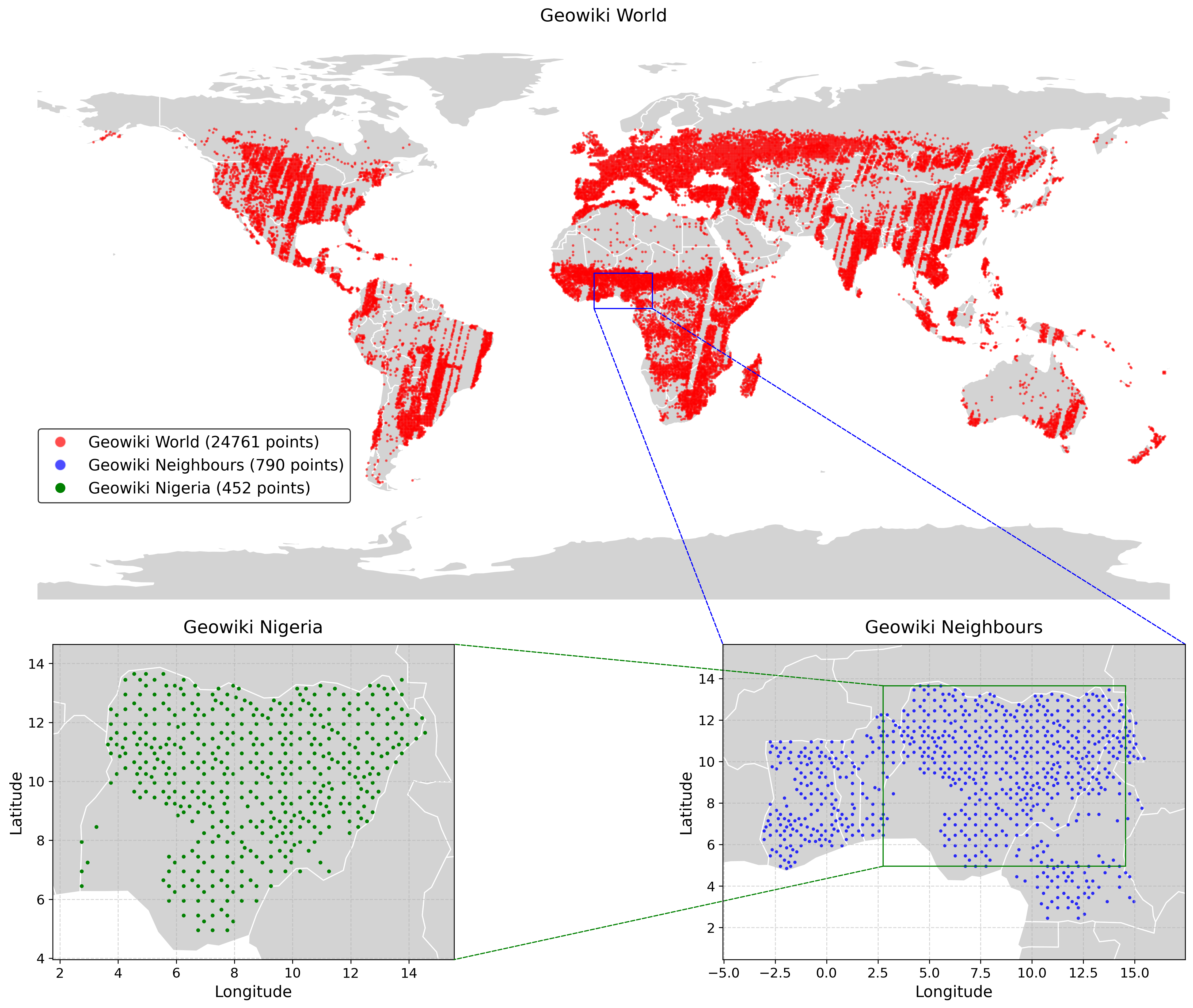}
  \caption{Distribution of crowd-sourced cropland labels from the Geowiki dataset \cite{geowiki_laso_bayas_global_2017}, and the different subsets used in this study: worldwide, Nigeria, and neighbouring countries. The neighbouring countries subset includes points in Nigeria, Ghana, Togo, Benin, and Cameroon.}
  \label{fig:geowiki_points}
\end{figure}

We first transformed the average Geowiki dataset labels into hard binary classifications (0 for non-cropland or 1 for cropland) using a crop probability threshold of 0.5. For our experiments, we then organized the data into three configurations, which are illustrated in Figure \ref{fig:geowiki_points}:

\begin{itemize}
    \item \textbf{Geowiki Nigeria:} contains all Geowiki samples within Nigeria's boundaries.
    \item \textbf{Geowiki Neighbours:} extends the Geowiki Nigeria dataset by including samples from neighbouring countries that share agro-ecological zones: Ghana, Togo, Benin, and Cameroon.
    \item \textbf{Geowiki World:} includes all Geowiki dataset samples available in the CropHarvest package.
\end{itemize}

We randomly split the Geowiki dataset configurations into 80 \% training and 20 \% validation data. In contrast, we split the Nigeria dataset into 50 \% train, 25 \% validation, and 25 \% test sets using a stratified random sampling strategy, ensuring that points in the validation and test sets are at least 30 km apart to avoid any possible data leakeage due to spatial autocorrelation~\cite{Li2023}. We chose this larger proportion of validation and test data to ensure a sizeable test set for a robust evaluation across Nigeria's diverse landscapes. The stratified random splits covering the whole country guarantee both the quality of the output map at identifying cropland across the different climate and vegetation types found in the country and representative test data. 
Table~\ref{tab:data_dist} summarises the final datasets and their class distributions across splits.

\begin{table}[htbp]
  \small
  \centering
  \caption{Class distribution across datasets and their splits. For the Geowiki datasets, only training and validation splits were created, while the Nigeria dataset also includes a test split which is used to compare our models. The cropland ratio refers to the proportion of points labelled as cropland.}
  \label{tab:data_dist}
  \begin{tabular}{@{}llccc@{}}
    \toprule
    Dataset & Split & Total points & Cropland points & Cropland ratio \\ \midrule
    \multirow{3}{*}{Geowiki Nigeria}
    & Train & 361 & 244 & 0.676 \\
    & Validation & 91 & 68 & 0.747 \\
    & Total & 452 & 312 & 0.690 \\ \midrule
    \multirow{3}{*}{Geowiki Neighbours}
    & Train & 632 & 364 & 0.576 \\
    & Validation & 158 & 96 & 0.608 \\
    & Total & 790 & 460 & 0.582 \\ \midrule
    \multirow{3}{*}{Geowiki World} 
    & Train & 19,808 & 11,131 & 0.562 \\
    & Validation & 4,953 & 2,849 & 0.575 \\
    & Total & 24,761 & 13,980 & 0.565 \\ \midrule
    \multirow{4}{*}{Nigeria}
    & Train & 913 & 381 & 0.417 \\
    & Validation & 454 & 181 & 0.399 \\
    & Test & 455 & 183 & 0.402 \\
    & Total & 1,822 & 745 & 0.409 \\ \bottomrule
  \end{tabular}
\end{table}

We experimented with various combinations of the Geowiki dataset configurations for training, both including and excluding the Nigeria dataset. For data normalization, we first calculated per-channel means and standard deviations independently using their respective training and validation samples, and aggregating over the time dimension. When merging datasets, following the approach in \cite{kerner_rapid_2020}, we combined all samples of the respective splits and calculated weighted per-channel means and standard deviations based on the relative number of samples from each dataset. We then applied per-channel z-score normalization to all samples across all data splits, i.e., each sample was normalized by subtracting the mean and then dividing by the standard deviation for each respective channel.

\subsection{Models}
\begin{figure}[htbp]
  \centering
  \includegraphics[scale=0.7]{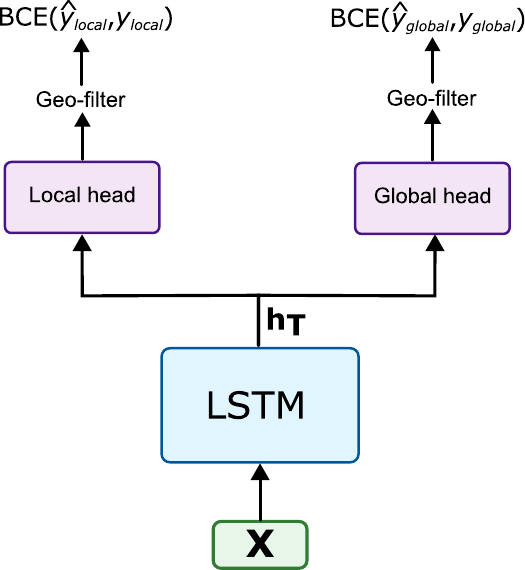}
  \caption{Illustration of the multi-head LSTM architecture for cropland classification introduced by \cite{kerner_rapid_2020}. A batch of input pixel time series arrays, $\mathbf{X} \in \mathbb{R}^{T \times C}$ ($T=12$ and $C=18$ in our study), is fed into a recurrent LSTM cell, which sequentially processes the time series and produces a final hidden state $\mathbf{h_T}$. The batch of hidden states is then passed to two separate MLPs (heads) and a final sigmoid activation function, yielding predictions bounded between 0 and 1. These predictions are then filtered based on the geographical location of the respective input points, with each head supervised exclusively on data from either inside or outside Nigeria using a Binary Cross Entropy loss (BCE). This multi-task learning design ensures that the local MLP specialises in classifying points within Nigeria while still benefiting from patterns learned by the LSTM through the global head supervision.}
  \label{fig:model}
\end{figure}

We followed \cite{kerner_rapid_2020,tseng_gabriel_annual_2020} and experimented with two LSTM neural network architectures: a standard one \cite{LSTM_HochSchm97} and a multi-task (multi-head) implementation \cite{kerner_rapid_2020}. LSTM networks are a type of RNN that excel in processing sequential data such as text or time series data and capturing long-term relationships. They have been shown to effectively capture complex temporal patterns in remote sensing data for crop classification \cite{russwurm_temporal_2017, turkoglu_crop_2021}, and they are parameter-efficient, which makes them well-suited for small-data regimes settings such as this work. We trained the models for binary classification of individual pixels. Each pixel in the input data represents a regularly sampled time series $\mathbf{X}=[\mathbf{x}_1, \mathbf{x}_2, ..., \mathbf{x}_{12}]$ of 12 time steps of 18 dimensions each, i.e. $\mathbf{x}_t \in \mathbb{R}^{1 \times 18}$. The model can have either one (single-headed) or two (multi-headed) identical classification heads consisting of a Multilayer Perceptron (MLP) and a sigmoid activation function to normalize the last activation between 0 and 1. This output value can be interpreted as a posterior probability of cropland presence. When using the multi-head architecture, both classification heads share the same LSTM backbone, but the local head is trained only with the samples that fall within Nigeria and the global head with all other samples. This architecture follows a multi-task learning approach \cite{tseng_learning_2021}, in which classifying global and local samples are treated as separate tasks handled by a dedicated head, but where relevant shared patterns are learned by the LSTM backbone. Once trained, the local head serves as a specialised classifier for the region of interest. A representation of the multi-head LSTM architecture is shown in Figure \ref{fig:model}.

We train the multi-head architecture by minimizing the loss function in Eq.~\eqref{eq:loss_function} \cite{kerner_rapid_2020}, where $\mathcal{L}_{local}$ and $\mathcal{L}_{global}$ are Binary Cross-Entropy (BCE) loss functions for the local and global classification heads respectively. $\frac{W}{\alpha}$ is a weighting term, where $W$ is the proportion of global to local labels in each batch and $\alpha$ is a weighting hyper-parameter. This weighting term reduces the impact of the global head in the overall loss function and thus its influence over the parameters updates of the shared LSTM backbone, effectively giving more relevance to the local classification task in the overall architecture. The single-head model uses one classifier for all samples regardless of their origin, treating them equally as "local" samples, i.e. with $W=0$ in the loss function.

\begin{equation}\label{eq:loss_function}
  \mathcal{L} = \frac{W}{\alpha}\mathcal{L}_{global} + \mathcal{L}_{local}
\end{equation}

\begin{equation}\label{eq:loss_function_weighted}
  \mathcal{L}_{global | local}  = - (w^{(1)} \cdot y \cdot log(\hat{y}) + w^{(0)} \cdot(1-y) \cdot log(1-\hat{y}))
\end{equation}

For this work, both $\mathcal{L}_{local}$ and $\mathcal{L}_{global}$ BCE loss terms were additionally weighted according to the class distributions, as in Eq.~\eqref{eq:loss_function_weighted}, to better account for class imbalance. In here, $w^{(1)}$ and $w^{(0)}$ respectively represent the inverse of the cropland and non-cropland instances proportion from the total amount of local or global labels, $y$ is the label and $\hat{y}$ is the model prediction. 

We kept all hyper-parameters to default values from \cite{kerner_rapid_2020}. These include one hidden layer size of 64 units for the LSTM backbone, two classification layers on the MLPs, dropout of 0.2 between LSTM state updates and an $\alpha$ of 10. We trained all models using the Adam optimizer \cite{kingma2017adam}, with a learning rate of 0.001, and a batch size of 64, for a maximum of 100 epochs and early stopping with patience of 10 epochs on the validation set loss. A grid-search of \{32, 64, 128\} hidden units and \{1, 2\} LSTM hidden layers on the validation set confirmed that these are robust hyper-parameters for our datasets. We implemented the DL models in PyTorch (version 1.13.1 with CUDA 11.7) \cite{Pytorch-NEURIPS2019_9015} and also included a RF baseline model from scikit-learn \cite{scikit-learn} (version 1.0.2) default implementation, due to its robustness and widespread use for crop and land cover classification \cite{PELLETIER2016156}. The model uses 100 trees with bootstrapping, where each tree is trained on random samples drawn with replacement from the full dataset. The complete source code was implemented in Python and is publicly available (see Data and code availability section).

\subsection{Evaluation metrics}
We calculated different performance metrics on the Nigeria test set to evaluate both our models and the land cover maps quantitatively. For the land cover maps, we considered the "crops" class as positive and any other prediction as negative. Our evaluation used precision (proportion of correct predictions), recall (proportion of correctly identified samples or "probability of detection"), F1-score (harmonic mean between precision and recall), total accuracy (global proportion of correctly classified samples), and the Area Under the Receiver Operating Characteristic curve (AUC ROC). These metrics are typically used in classification settings and AUC ROC is exclusively used in binary classification for evaluating the ability of a classifier to correctly distinguish between the positive and negative classes \cite{sokolova_systematic_2009, fawcett_introduction_2006}. For all metrics a higher value is better, however high precision often comes at the cost of lower recall and vice versa. The F1-score captures this trade-off by providing a balanced single metric. Eqs.~\eqref{eq:metrics} presents the equations for precision, recall, F1-score, total accuracy, and False Positive Rate (FPR), where TP, FP, FN, and TN are true positives, false positives, false negatives, and true negatives predictions, respectively \cite{sokolova_systematic_2009}.

To calculate precision, recall, F1-score and total accuracy, we needed to convert model prediction probabilities into hard binary predictions (0 for non-cropland or 1 for cropland). We used a threshold of 0.5, as in \cite{kerner_rapid_2020}, which is a sensible choice given there is no clear preference for a higher True Positive Rate (TPR) or FPR in our cropland mapping use case. This threshold is only required for model performance evaluation and is not used during model training. For the AUC ROC score, the TPR (computed the same as recall) and the FPR (or "probability of false alarm") at different thresholds values are calculated to build a ROC curve and the area under the curve is computed \cite{fawcett_introduction_2006}. The scikit-learn Python package \cite{scikit-learn} was used for calculating all metrics.

\begin{gather}
  \label{eq:metrics}
  \begin{aligned}
    \text{Precision} &= \frac{\text{TP}}{\text{TP} + \text{FP}} \\
    \text{Recall} &= \frac{\text{TP}}{\text{TP} + \text{FN}} \\
    \text{F1-score} &= \frac{2 \times \text{Precision} \times \text{Recall}}{\text{Precision} + \text{Recall}} \\
    \text{Total Accuracy} &= \frac{\text{TP + \text{TN}}}{\text{TP + FP + FN + TN}} \\
    \text{FPR} &= \frac{\text{FP}}{\text{FP} + \text{TN}}
  \end{aligned}
\end{gather}

\section{Results}
\subsection{Main experiments}
In this section we present the experimental results of the different models on the Nigeria test set, considering the different dataset configurations discussed in Section \ref{sec:dataset_preprocessing}. The results are summarised in Table \ref{tab:weighted_main_results}.

\begin{table}[htbp]
\small 
\centering
\caption{Performance comparison of different models and land cover maps on the Nigeria test set. In the dataset columns, \checkmark indicates the model was trained using that dataset, $\times$ indicates training without that dataset, and `-' indicates not applicable for map products (AUC ROC due to hard predictions). As described in Section \ref{sec:dataset_preprocessing}, the Geowiki Neighbours subset of the Geowiki dataset \cite{geowiki_laso_bayas_global_2017}, includes points from Nigeria, Ghana, Togo, Benin, and Cameroon. We computed the metrics by applying a 0.5 threshold on model predictions. Best results per metric are in \textbf{bold}, and second best are \underline{underlined}.}
\begin{tabular}{@{}c@{\hspace{8pt}}cc@{\hspace{16pt}}ccccc@{}}
\toprule[1.2pt]
\multicolumn{3}{c}{Model Configuration} & \multicolumn{5}{c}{Performance Metrics} \\ 
\cmidrule[1.2pt](r{1.5em}){1-3} \cmidrule[1.2pt]{4-8}
Map / Model & Geowiki dataset & Nigeria dataset & AUC ROC & Precision & Recall & F1-score & Accuracy \\ 
\cmidrule[1.2pt](r{1.5em}){1-3} \cmidrule[1.2pt]{4-8}
ESA WorldCover 2020  & -       & -       & -  & \underline{0.903} & 0.760 & \textbf{0.825} & \textbf{0.870} \\
ESRI 2020           & -       & -       & -  & 0.867 & 0.213 & 0.342 & 0.670 \\
Dynamic World 2020  & -       & -       & -  & 0.491 & 0.448 & 0.469 & 0.591 \\ 
\cmidrule[1pt](r{1.5em}){1-3} \cmidrule[1pt]{4-8}
                    & $\times$ & \checkmark & 0.911 & 0.774 & 0.787 & 0.780 & 0.822 \\
                    & Nigeria & \checkmark & \underline{0.915} & 0.782 & 0.825 & 0.803 & 0.837 \\
                    & Neighbours & \checkmark & \textbf{0.916} & 0.791 & 0.765 & 0.778 & 0.824 \\
Random Forest       & World & \checkmark & 0.796 & 0.569 & 0.858 & 0.684 & 0.681 \\
                    & Nigeria & $\times$ & 0.755 & 0.495 & \underline{0.891} & 0.637 & 0.591 \\
                    & Neighbours & $\times$ & 0.842 & 0.695 & 0.798 & 0.743 & 0.778 \\
                    & World & $\times$ & 0.755 & 0.522 & 0.836 & 0.643 & 0.626 \\ 
\cmidrule[0.5pt](r{1.5em}){1-3} \cmidrule[0.5pt]{4-8}
                    & $\times$ & \checkmark & 0.908 & 0.772 & 0.776 & 0.774 & 0.818 \\
                    & Nigeria & \checkmark & 0.907 & 0.771 & 0.863 & \underline{0.814} & \underline{0.842} \\
                    & Neighbours & \checkmark & 0.906 & 0.788 & 0.831 & 0.809 & \underline{0.842} \\
Single-headed LSTM  & World & \checkmark & 0.884 & 0.671 & 0.847 & 0.749 & 0.771 \\
                    & Nigeria & $\times$ & 0.756 & 0.561 & 0.749 & 0.642 & 0.664 \\
                    & Neighbours & $\times$ & 0.874 & 0.688 & 0.831 & 0.752 & 0.780 \\
                    & World & $\times$ & 0.854 & 0.575 & \textbf{0.902} & 0.702 & 0.692 \\ 
\cmidrule[0.5pt](r{1.5em}){1-3} \cmidrule[0.5pt]{4-8}
                    & Neighbours & \checkmark & 0.889 & 0.737 & 0.874 & 0.800 & 0.824 \\ 
                    & World & \checkmark & 0.893 & \textbf{0.917} & 0.541 & 0.680 & 0.796 \\ 
Multi-headed LSTM   & Neighbours & $\times$ & 0.772 & 0.638 & 0.694 & 0.665 & 0.719 \\
                    & World & $\times$ & 0.819 & 0.721 & 0.650 & 0.684 & 0.758 \\ 
\bottomrule[1.2pt]
\end{tabular}
\label{tab:weighted_main_results}
\end{table}

Among the implemented models, the single-headed LSTM model achieved the best performance when trained with the Nigeria dataset combined with either the Geowiki Nigeria or Geowiki Neighbours subset, both reaching an accuracy of 0.842. The Geowiki Nigeria configuration slightly outperformed the Neighbours configuration with an F1-score of 0.814 compared to 0.809, primarily due to its higher recall (0.863 vs 0.831).

The addition of the Nigeria dataset consistently improved model performance across all model architectures. For example, when using the Geowiki Nigeria dataset, the single-headed LSTM's accuracy improved from 0.664 to 0.842 with the addition of the Nigeria dataset, and similarly, the RF model showed a dramatic improvement from 0.591 to 0.837. When trained without the Nigeria dataset, using the Geowiki Neighbours subset improved performance compared to using only the Geowiki Nigeria subset, as seen in terms of accuracy and F1-score in the RF (0.778 vs 0.591, and 0.743 vs 0.637, respectively) and in the single-headed LSTM models (0.780 vs 0.664, and 0.752 vs 0.642, respectively). However, using the full Geowiki World dataset generally led to decreased performance, except for the multi-headed LSTM which showed improvements in accuracy (0.719 vs 0.758), F1-score (0.665 vs 0.684) and AUC ROC (0.772 vs 0.819). Additionally, when using only the Nigeria dataset without any Geowiki data, the RF model slightly outperformed the single-headed LSTM, achieving higher accuracy (0.822 vs 0.818) and F1-score (0.780 vs 0.774). 

Despite these variations across dataset configurations, the multi-headed LSTM model showed comparable or better performance than the RF model under similar configurations. When trained with the Nigeria dataset and Geowiki Neighbours subset, both models achieved the same accuracy (0.824), but the multi-headed LSTM achieved a higher F1-score (0.800 vs 0.778). Similarly, when using Geowiki World without the Nigeria dataset, the multi-headed LSTM outperformed the RF in accuracy (0.758 vs 0.626), F1-score (0.684 vs 0.643), and AUC ROC score (0.819 vs 0.755).

\begin{figure}[htbp]
\centering
  \includegraphics[scale=0.5]{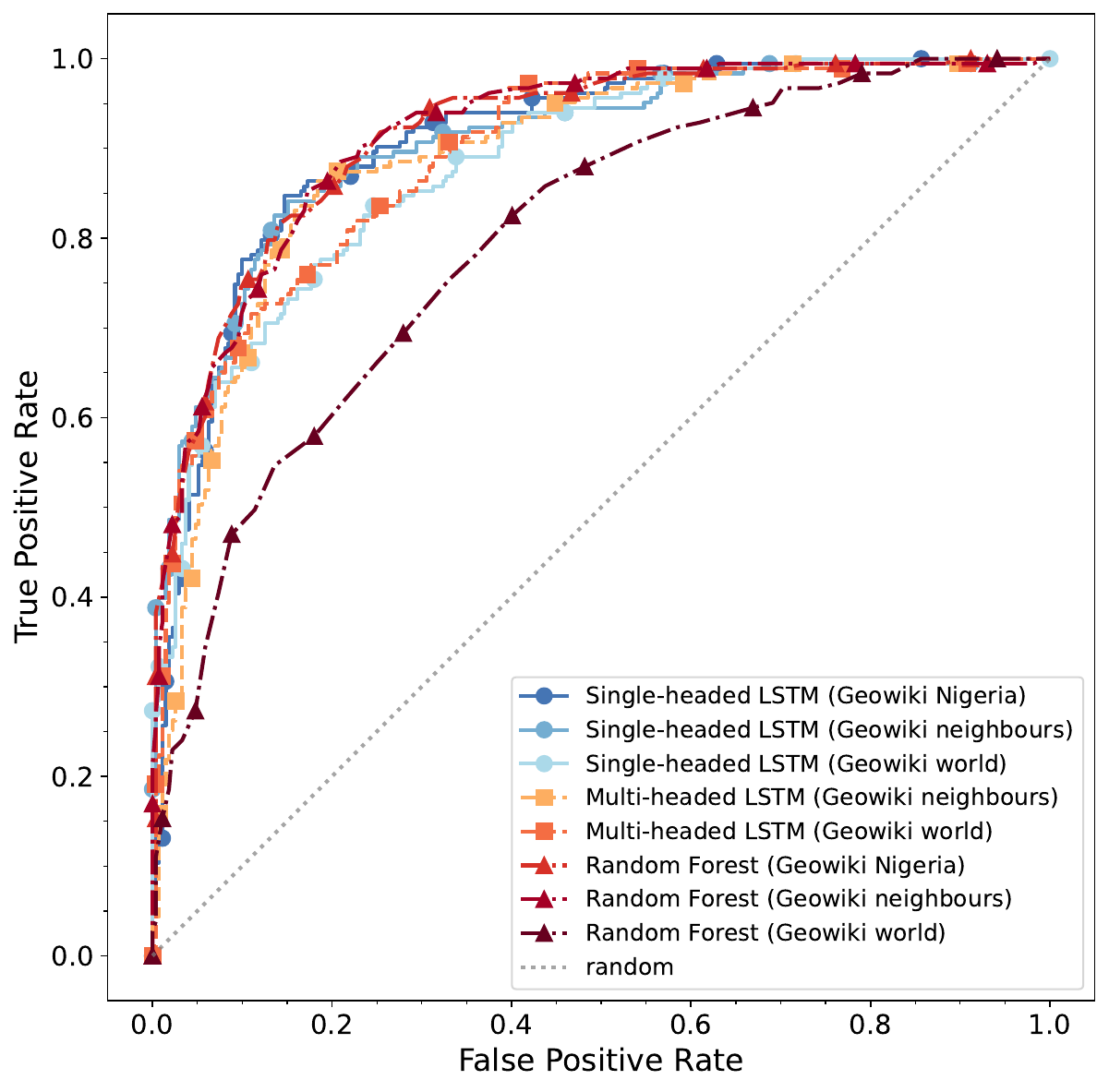}
  \caption{Receiver Operating Characteristic (ROC) curves on the Nigeria test set. All models represented here are trained with the Nigeria dataset and the Geowiki subset specified in the legend. The 45° line is equivalent to a random prediction (0.5 in expectation) at different thresholds. The closer the curve is to the upper left corner, the better the classifier.}
  \label{fig:roc_curve_models}
\end{figure}

Furthermore, the AUC ROC metric offers insights into model performance under different threshold values, and in Figure \ref{fig:roc_curve_models} we present the ROC curves for all models using the Nigeria dataset and Geowiki data. Most models showed similar ROC curves, with the RF model achieving higher TPR for FPR between 0.2 and 0.6, corresponding to increasingly lower thresholds. Conversely, models using the Geowiki World dataset showed consistently lower performance across thresholds, with the RF model experiencing the most substantial decrease.

When comparing our models to existing global land cover maps as benchmarks, we found that the ESA WorldCover 2020 map achieved the highest F1-score (0.825) and total accuracy (0.870). However, its recall of 0.760 was lower than our best models, such as the single-headed LSTM with Nigeria datasets (0.863) and the multi-headed LSTM with the Geowiki Neighbours and the Nigeria dataset (0.874), while maintaining higher precision (0.903 vs 0.771 and 0.737). The other global maps, ESRI 2020 and Dynamic World 2020, achieved substantially lower performance metrics compared to our implemented models.

\subsection{Sensitivity analysis}
We conducted additional experiments to evaluate model sensitivity to input features and the weighted BCE loss function. Using only Sentinel-2 bands and NDVI (Table S1) led to consistent performance degradation across all models compared to using the full feature set (Table \ref{tab:features}). For example, when using the Nigeria and Geowiki Neighbours datasets, the RF model's accuracy decreased from 0.824 to 0.804 and F1-score from 0.778 to 0.753. Meanwhile, the single-headed LSTM experienced the largest decrease in this configuration, with accuracy dropping from 0.842 to 0.774 and F1-score from 0.809 to 0.710. However, the most substantial performance reduction overall occurred in the multi-headed LSTM model under the Geowiki World without the Nigeria dataset configuration, with accuracy dropping from 0.758 to 0.604 and F1-score from 0.684 to a mere 0.091. In contrast, when additionally trained with the Nigeria dataset, the same model only dropped its performance from 0.796 to 0.771 in accuracy, and from 0.680 to 0.679 in F1-score.

For the LSTM models, we additionally evaluated the impact of using a regular BCE loss instead of the weighted version (Table S2). With the Nigeria and Geowiki Neighbours datasets, the single-headed LSTM showed minimal changes in accuracy (0.842 vs 0.837) and F1-score (0.809 vs 0.800). The multi-headed LSTM exhibited similarly stable performance, with accuracy increasing slightly from 0.824 to 0.826 and F1-score from 0.800 to 0.801. However, larger differences emerged when using only Geowiki data. The single-headed LSTM's accuracy dropped from 0.780 to 0.752, and its F1-score decreased from 0.752 to 0.748 in the Geowiki Neighbours configuration. Additionally, when using only the Geowiki Nigeria dataset, the single-headed LSTM's accuracy fell from 0.664 to 0.593, and its F1-score from 0.642 to 0.633. Instead, when using only the Nigeria dataset, the same model maintained its performance, showing a minor improvement in accuracy (+0.02), with a slight decrease in F1-score (-0.02) and AUC ROC (-0.01).

Additionally, we computed performance metrics for all three land cover maps on the Togo dataset (Section \ref{sec:togo_dataset}) and included other published results for comparison (Table \ref{tab:other_countries}). The multi-headed LSTM model from \cite{kerner_rapid_2020} achieved the highest metrics (F1-score: 0.74, accuracy: 0.83), while among the land cover maps, Dynamic World achieved the highest F1-score (0.678) and WorldCover the highest accuracy (0.810) and precision (0.875), but with a recall of only 0.528.

\subsection{Nigeria cropland maps}
\begin{figure}[htbp]
\centering
\includegraphics[clip, trim=1cm 5cm 1cm 5cm, scale=0.35]{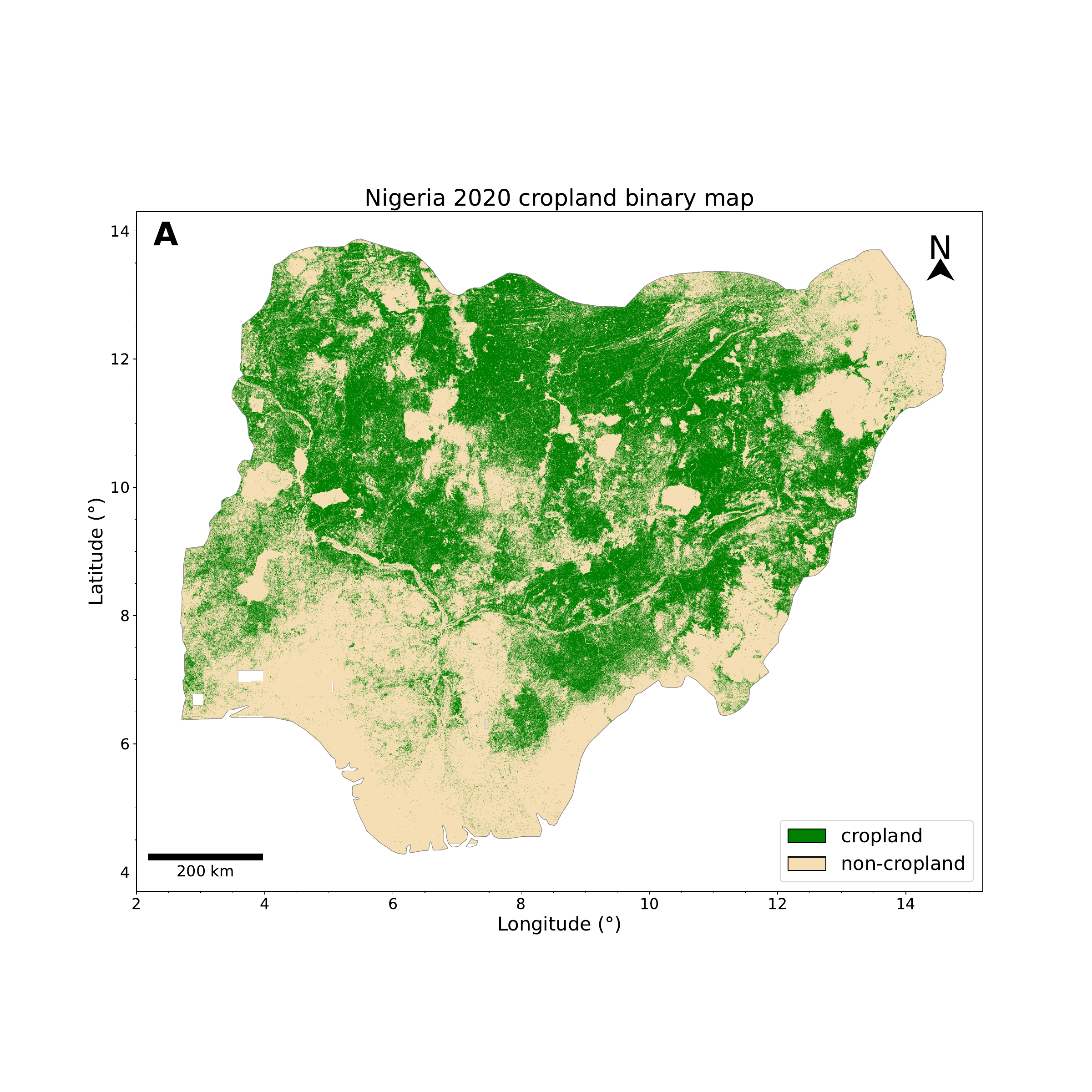}
\\
\vspace{-0.3cm} 
\includegraphics[clip, trim=1cm 5cm 1cm 5cm, scale=0.35]{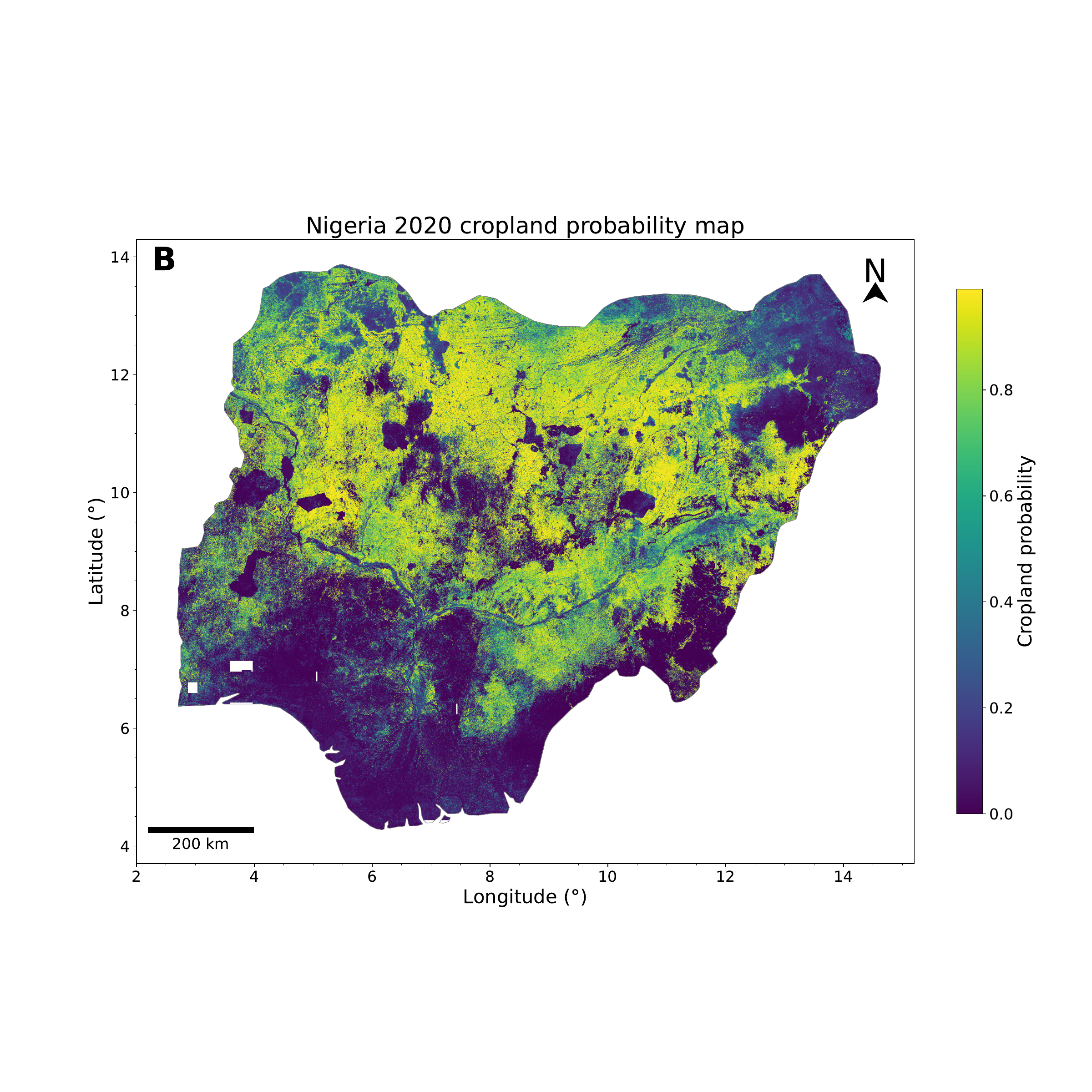}
\caption{Binary (with a threshold of 0.5) and probability cropland maps for Nigeria in 2020 (A and B, respectively), generated with our best model: a single-headed LSTM model trained with the Geowiki Nigeria and the Nigeria datasets. The few small white patches in the southwest region are regions where input satellite data was unavailable.}
\label{fig:final_maps}
\end{figure}

\begin{figure}[htbp]
\centering
  \includegraphics[width=0.95\textwidth]{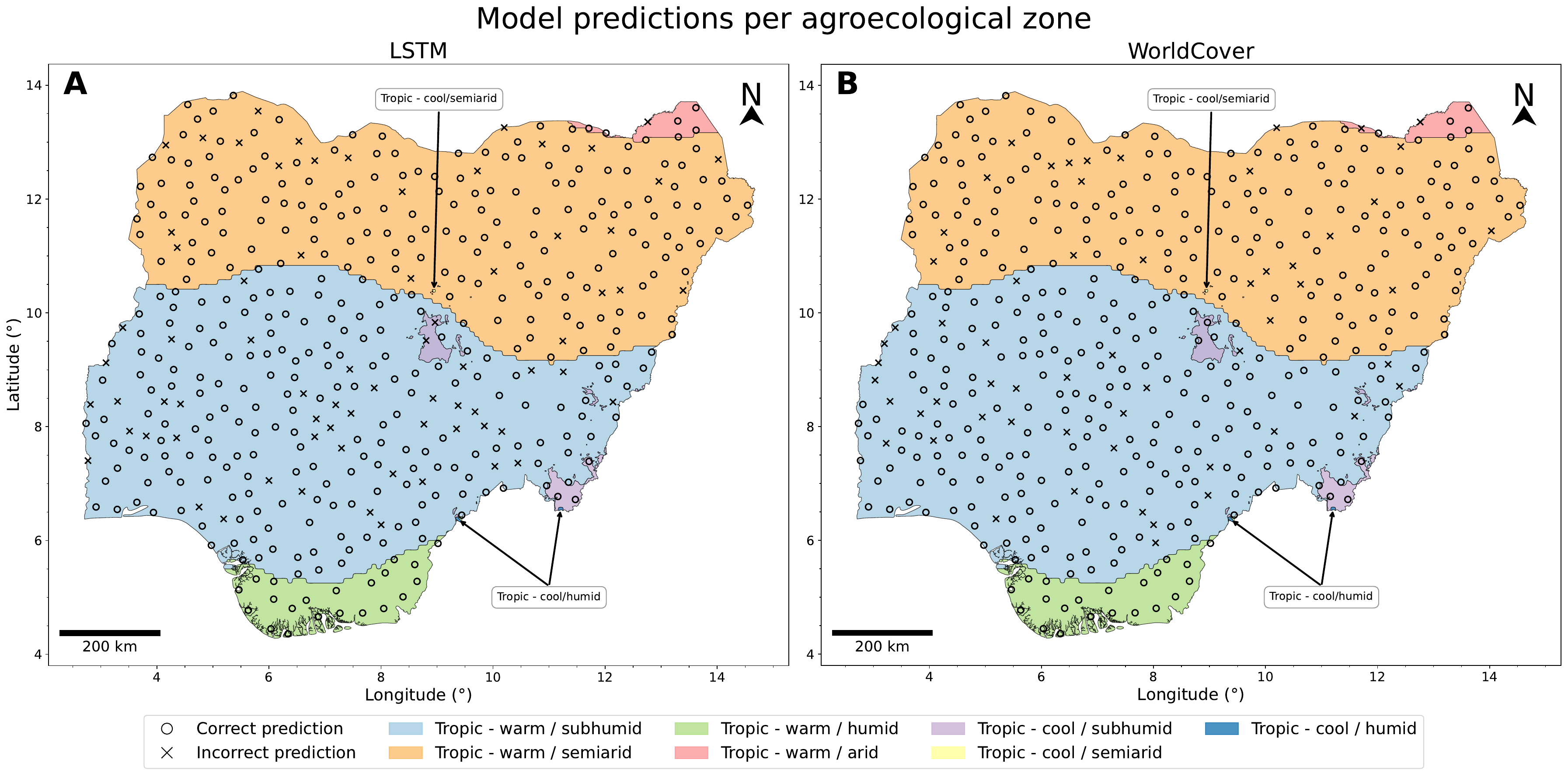}
  \caption{Predictions of the best model (A) and WorldCover (B) on the Nigeria test set by agroecological zone \cite{agrozones-nigeria-sambus}.}
  \label{fig:agrozones}
\end{figure}

We used our best model, the single-headed LSTM trained with the Geowiki Nigeria subset and the Nigeria dataset, to generate both binary and probabilistic cropland maps covering the entire extent of Nigeria for the year 2020. 
The maps are presented in Figure \ref{fig:final_maps}, with a few small white patches in the southwest region indicating areas where input satellite data was unavailable. We refer the reader to our supplementary GEE web app for an interactive visualization of the map (link in Data and code availability section). In addition, we present the results of the single-headed LSTM model and the WorldCover map aggregated by agroecological zone \cite{agrozones-nigeria-sambus} in Table \ref{tab:agrozone} and Figure \ref{fig:agrozones}. Excluding the agroecological zones represented with just a few points, the \textit{Tropic - warm / subhumid} zone accounts for the largest performance difference between the model and the land cover map (0.806 vs 0.853 accuracy).

\begin{table}[htbp]
\small
\centering
\caption{Results of the best model, the single-headed LSTM, and the ESA WorldCover 2020 land cover map on the Nigeria test set by agroecological zone.}
\label{tab:agrozone}
\begin{tabular}{lcccc}
\toprule
Agroecological Zone  & Total points & Total area ($km^2$) & LSTM accuracy & WorldCover accuracy \\
\midrule
Tropic - warm / subhumid & 232       & 462,634            & 0.806                  & 0.853        \\
Tropic - warm / semiarid & 191       & 420,257            & 0.864                  & 0.880        \\
Tropic - warm / humid    & 19        & 39,901             & 1.0                    & 1.0          \\
Tropic - warm / arid     & 5         & 8,398              & 1.0                    & 0.8          \\ 
Tropic - cool / subhumid & 4         & 18,712             & 0.75                   & 1.0          \\ 
\bottomrule
\end{tabular}
\end{table}

\section{Discussion}
\subsection{Analysis}

The impact of dataset composition on model performance revealed interesting patterns across architectures. The addition of the Nigeria dataset consistently improved performance across all models, demonstrating the importance of region-specific training data. However, when local data was unavailable (experiments without the Nigeria dataset), using the Geowiki Neighbours subset proved more effective than the complete Geowiki World dataset for most models. This suggests that geographical proximity of training samples may be more valuable than larger but more geographically diverse datasets, possibly due to more similar agricultural practices and environmental conditions in neighbouring regions. The multi-headed LSTM was a notable exception to this pattern, showing improved performance with the full Geowiki World dataset, indicating that certain architectural choices may better facilitate knowledge transfer from diverse geographical contexts \cite{tseng2022timl, tseng_lightweight_2024}. These findings align with recent work by \cite{kerner2023accurate}, who found that models optimized for regional performance regularly outperformed globally trained models. Similarly, \cite{Hacheme2024} demonstrated that supplementing sparse but targeted local annotations with weak labels—even when derived from the same region—can degrade classification performance. \cite{Weitkamp2023} further highlighted that including locally relevant training data substantially improved the accuracy of irrigated agriculture mapping in Mozambique, especially in heterogeneous smallholder systems.

Comparing the DL approaches with the RF baseline revealed complementary strengths. While RF models performed well with locally focused training data, they were more sensitive to geographically diverse inputs, showing greater performance degradation on the full Geowiki dataset. In contrast, DL models demonstrated greater robustness to data variability, likely due to their ability to capture complex spatio-temporal patterns. This supports a broader trend in the field towards large-scale, self-supervised DL architectures trained on global remote sensing datasets, which can have shown strong generalization across regions and tasks \cite{tseng_lightweight_2024,Tseng2025,Astruc2024,Nedungadi2024}. While these foundation models offer strong generalization potential, our study focused on evaluating performance in fully supervised settings using models trained specifically on regionally tailored data. A methodical evaluation of foundation models using different probing or fine-tuning strategies in Nigeria was therefore beyond the scope of this study and is left for future work.

Our sensitivity analysis highlighted the importance of additional features beyond Sentinel-2 bands and NDVI. The removal of supplementary features (SAR, climate and topological features from Table \ref{tab:features}) led to consistent performance degradation across almost all models, with the LSTM models showing greater sensitivity to feature reduction compared to RF models. Most notably, when training the multi-headed LSTM only with Geowiki World data, removing the additional features caused a dramatic drop in F1-score from 0.684 to 0.091, while the same model maintained relatively stable performance when the Nigeria dataset was included in training. An exception was the single-headed LSTM model trained only with the Geowiki World data, where its accuracy increased from 0.692 to 0.721 and its F1-score from 0.702 to 0.713, but its AUC ROC decreased from 0.854 to 0.804. This suggests that these additional features are important for distinguishing cropland patterns, particularly when learning from geographically diverse data and when using more complex architectures \cite{adrian_sentinel_2021}. Sentinel-1 time series and environmental data, have also been found by other studies to be particularly useful for crop type mapping in large regions \cite{reus_comparison_2021, blickensdorfer_mapping_2022}. Our results also align with a growing body of work on multimodal learning in remote sensing, which shows that integrating complementary data types—such as SAR, optical, and climate information—substantially improves model robustness and generalization \cite{Hong2020}. Across both supervised models and recent foundation models, the question of how to most effectively fuse these modalities remains an active area of research \cite{Tseng2025, Mena2025}.

Furthermore, our analysis on the weighted loss function revealed insightful patterns regarding the handling of class imbalance in cropland detection. When the training and test datasets had similar cropland ratios, such as when using the Nigeria dataset (train: 0.417, test: 0.402), the choice between weighted and regular BCE loss had minimal impact on model performance. Likewise, when combining the Nigeria dataset with Geowiki Neighbours data (combined cropland ratio of 0.482), both LSTM architectures exhibited stable performance regardless of the loss function. However, the importance of loss weighting became more pronounced when training exclusively with Geowiki data, where the cropland ratios differ significantly from the Nigeria test set. This discrepancy was particularly evident when using the Geowiki Nigeria dataset, where the cropland ratio of the train set (0.690) was substantially higher than that of the Nigeria test set (0.402). In this scenario, the accuracy and F1-score of the single-headed LSTM dropped from 0.664 to 0.593 and 0.642 to 0.633 respectively, when switching to a regular BCE loss. These findings suggest that appropriate handling of class imbalance becomes crucial when there are significant differences in class distributions between the training and test datasets. Alternatively, this challenge can be mitigated by increasing the representation of minority classes, for example by mining weak labels from coarse-resolution global products \cite{Hacheme2024}, or by generating synthetic samples \cite{Mirzaei2023}.

Finally, the performance of our best model (accuracy: 0.842, F1-score: 0.814, recall: 0.863) is comparable to similar approaches in other African countries, such as cropland mapping in Kenya (accuracy: 0.86, F1-score: 0.84) \cite{tseng_gabriel_annual_2020} and Togo (accuracy: 0.83, F1-score: 0.74) \cite{kerner_rapid_2020}, as well as a study in the Jos region of Nigeria (accuracy: 0.84) \cite{Ibrahim2021}. When compared to global land cover products, the ESA WorldCover map performed well on the Nigeria test set (accuracy: 0.87, F1-score, 0.825, recall: 0.760), particularly in the Tropic-warm/subhumid region. This may be due to its design features, as previous studies have highlighted WorldCover's lower minimum mapping unit (100 m²) compared to ESRI Land Cover and Dynamic World as a key factor for its high accuracy and spatial detail, which makes it well-suited for heterogeneous regions such as smallholder farming landscapes \cite{venter_global_2022, xu_comparative_2024}. Another contributing factor may be its inclusion of Sentinel-1 data, which we have found to be a valuable input for cropland mapping in Nigeria. Nevertheless, WorldCover's limitations remain evident, including its lower performance in Togo (Table S3) and a tendency towards lower recall overall \cite{kerner2023accurate}, making it prone to underestimating cropland areas, which is particularly concerning in regions facing food security risks exacerbated by climate change. This is consistent with findings by \cite{Tadesse2024}, who observed poor performance of WorldCover in Kenya and found that a locally trained model substantially outperformed the global product, underscoring the importance of localized models for accurate cropland monitoring in smallholder-dominated systems

High spatial accuracy and reliability in these maps is critical not only for land cover monitoring but also for downstream applications such as crop yield modeling and forecasting \cite{Mena2025}, or estimating irrigation demand and water use in agricultural systems \cite{Weitkamp2023}. As such, the development of localized, robust cropland maps can directly support decision-making in agricultural policy, food security planning, and climate adaptation, especially in predominantly agrarian economies such as those accross sub-Saharian Africa \cite{Ibrahim2021,Tadesse2024}.

\subsection{Limitations}
While this study provides a comprehensive analysis of model architectures, dataset configurations, and feature relevance for country-scale cropland mapping, there are important limitations that should be acknowledged. First, the harmonization of class definitions across datasets remains an open challenge \cite{tseng2021cropharvest}. Our Nigeria dataset considers permanent crops as orchards and fruit trees as cropland, as defined by the Food and Agriculture Organization (FAO), given their importance for food security, while the Geowiki dataset does not consider perennial crops or fallow land as cropland \cite{geowiki_laso_bayas_global_2017}. Similarly, global land cover maps regard orchards and fruit trees within their tree cover class \cite{venter_global_2022}. These inconsistencies in class definitions could affect both model training and performance evaluation. Second, the new Nigeria dataset was created by photo-interpretation by a single annotator. While this approach allowed for consistent labelling criteria, it may introduce systematic biases, particularly in complex smallholder farming landscapes where land use interpretation can be challenging. Finally, we did not evaluate recent foundation models or advanced multimodal fusion strategies, which are becoming increasingly important in remote sensing. Exploring how such models perform in data-scarce African contexts is an important area for future research.


\section{Conclusion}
Accurate and timely cropland maps are essential for agricultural monitoring, policy making, and assessing climate change impacts on food security, particularly in regions where agricultural data is scarce. 
In this study, we systematically evaluated how the geographical composition of training data, across different machine and deep learning architectures, affects cropland mapping performance in Nigeria, a data-scarce and agriculturally important region. While the impact of input data modalities has been previously explored, the role of training data origin—and how global, regional, and local sources interact to influence model performance—has received less attention, particularly in smallholder-dominated landscapes.
Our results demonstrate that incorporating locally labeled data significantly improves model performance, with our best-performing model, a single-headed LSTM, achieving an F1-score of 0.814 when trained on a combination of our Nigeria dataset and the Nigeria subset of the global Geowiki dataset. Our analysis revealed three key insights: (1) regionally specific training data substantially improves model performance, likely due to better alignment with local cropping systems and environmental conditions. Conversely, we found that supplementing local data with globally distributed training samples (e.g., the full Geowiki World dataset) often degraded performance, while proximity-based regional subsets (e.g., Geowiki Neighbours) offered a more effective compromise.
(2) Multi-modal data integration is critical, as Sentinel-1, climate, and topographic data significantly enhanced model robustness, with their removal causing accuracy and F1-score drops of up to 0.154 and 0.593, respectively. (3) Proper class imbalance handling is essential when the class distribution of the training and test sets significantly differ, with weighted loss functions improving accuracy by up to 0.071 for the single-headed LSTM. Comparison with existing global land cover products showed that our models, particularly the single-headed LSTM, performed competitively with ESA WorldCover despite the modest dataset size, demonstrating superior recall, a key metric for food security applications. The open-source nature of our models and methodology enables their immediate application for food security monitoring, climate adaptation planning, and evidence-based agricultural policy-making, particularly in data-scarce regions. 
Future research should build on these findings by evaluating the generalization capacity of foundation models, exploring scalable fusion strategies, and addressing class harmonization challenges. Further validation in real-world applications—such as yield estimation or irrigation monitoring—will help ensure that cropland mapping models contribute effectively to food security and climate resilience. 

\section*{Data and code availability}
\label{sec:data_and_code}
The source code, as well as the links to download the data and output maps, are publicly available in the following repository: \url{https://github.com/Joaquin-Gajardo/nigeria-crop-mask}. The maps can be visualized interactively in the following Google Earth Engine web application: \url{https://joaquingajardocastillo.users.earthengine.app/view/nigeria-cropland-maps}

\section*{Author contributions}
\textbf{Joaquin Gajardo}: Conceptualization, Investigation, Methodology, Software, Formal analysis, Data curation, Validation, Visualization, Writing- Original draft preparation, Writing- Reviewing and Editing. \textbf{Michele Volpi}: Supervision, Writing- Reviewing and Editing. \textbf{Daniel Onwude}: Project administration, Funding acquisition, Writing- Reviewing and Editing. \textbf{Thijs Defraeye}: Supervision, Project administration, Funding acquisition, Writing- Reviewing and Editing.

\section*{Acknowledgments}
This work was partially funded by the \url{data.org} Inclusive Growth and Recovery Challenge grant “Your Virtual Cold Chain Assistant”, supported by The Rockefeller Foundation and the Mastercard Center for Inclusive Growth, as well as by the project “Scaling up Your Virtual Cold Chain Assistant” commissioned by the German Federal Ministry for Economic Cooperation and Development and being implemented by BASE and Empa on behalf of the German Agency for International Cooperation (GIZ). The funders were not involved in the study design, collection, analysis, interpretation of data, the writing of this article, or the decision to submit it for publication.

\bibliographystyle{unsrt}  
\bibliography{references}  

\newpage

\section{Supplementary Material}
\beginsupplement

\begin{table}[hbtp]
\small
\centering
\label{tab:results_S2_only}
\caption{Results comparison on the Nigeria test set with only using Sentinel-2 bands and NDVI for training. Metrics were computed by applying a 0.5 threshold on model predictions. Best results per metric are in \textbf{bold} and second best are \underline{underlined}.}
\begin{tabular}{@{}cccccccc@{}}
\toprule
Model               & Geowiki dataset & Nigeria dataset  & AUC ROC & Precision & Recall & F1-score & Accuracy \\ \midrule
                    & $\times$        & \checkmark       &  0.880  &  0.749  &  0.765  &  0.757  &  0.802 \\
                    & Nigeria         & \checkmark       &  \textbf{0.887}  &  0.755  &  0.809  &  \textbf{0.781}  &  \textbf{0.818} \\
                    & Neighbours      & \checkmark       &  \textbf{0.887}  &  0.764  &  0.743  &  0.753  &  \underline{0.804} \\
Random Forest       & World           & \checkmark       &  0.722  &  0.587  &  0.667  &  0.624  &  0.677 \\
                    & Nigeria         & $\times$         &  0.606  &  0.471  &  0.574  &  0.517  &  0.569 \\
                    & Neighbours      & $\times$         &  0.778  &  0.723  &  0.470  &  0.570  &  0.714 \\
                    & World           & $\times$         &  0.691  &  0.530  &  0.732  &  0.615  &  0.631 \\ \midrule
                    & $\times$        & \checkmark       &  0.886  &  0.730  &  \underline{0.814}  &  \underline{0.770}  &  \underline{0.804} \\
                    & Nigeria         & \checkmark       &  0.869  &  \underline{0.781}  &  0.645  &  0.707  &  0.785 \\
                    & Neighbours      & \checkmark       &  0.870  &  0.733  &  0.689  &  0.710  &  0.774 \\
Single-headed LSTM  & World           & \checkmark       &  0.750  &  0.584  &  0.683  &  0.630  &  0.677 \\
                    & Nigeria         & $\times$         &  0.658  &  0.667  &  0.284  &  0.398  &  0.655 \\
                    & Neighbours      & $\times$         &  0.810  &  0.681  &  0.607  &  0.642  &  0.727 \\
                    & World           & $\times$         &  0.804  &  0.608  &  \textbf{0.863}  &  0.713  &  0.721 \\ \midrule
                    & Neighbours      & \checkmark       &  \underline{0.884}  &  \textbf{0.810}  &  0.650  &  0.721  &  0.798 \\ 
Multi-headed LSTM   & World           & \checkmark       &  0.847  &  0.780  &  0.601  &  0.679  &  0.771 \\ 
                    & Neighbours      & $\times$         &  0.808  &  0.702  &  0.475  &  0.567  &  0.708 \\
                    & World           & $\times$         &  0.600  &  0.600  &  0.050  &  0.091  &  0.604 \\  \bottomrule
\end{tabular}
\end{table}

\begin{table}[hbtp]
\small
\centering
\label{tab:main_results_not_weighted}
\caption{Results of LSTM model on the Nigeria test set without using a weighted BCE loss. Metrics were computed by applying a 0.5 threshold on model predictions. Best results per metric are in \textbf{bold} and second best are \underline{underlined}.}

\begin{tabular}{@{}cccccccc@{}}
\toprule
Model               & Geowiki dataset & Nigeria dataset  & AUC ROC & Precision & Recall & F1-score & Accuracy \\ \midrule
                    & $\times$        & \checkmark            &  \textbf{0.907}  &  \underline{0.797}  &  0.749  &  0.772  &  0.822  \\
                    & Nigeria         & \checkmark            &  \textbf{0.907}  &  0.764  &  0.847  &  \textbf{0.803}  &  \underline{0.833}  \\
                    & Neighbours      & \checkmark            &  \underline{0.905}  &  0.791  &  0.809  &  0.800  &  \textbf{0.837}  \\
Single-headed LSTM  & World           & \checkmark            &  0.868  &  0.704  &  0.792  &  0.746  &  0.782  \\
                    & Nigeria         & $\times$              &  0.785  &  0.497  &  0.874  &  0.633  &  0.593  \\
                    & Neighbours      & $\times$              &  0.872  &  0.632  &  \underline{0.918}  &  0.748  &  0.752  \\
                    & World           & $\times$              &  0.846  &  0.538  &  \textbf{0.929}  &  0.681  &  0.651  \\ \midrule
                    & Neighbours      & \checkmark            &  0.890  &  0.743  &  0.869  &  \underline{0.801}  &  0.826  \\
Multi-headed LSTM   & World           & \checkmark            &  0.892  &  \textbf{0.920}  &  0.437  &  0.593  &  0.758  \\
                    & Neighbours      & $\times$              &  0.827  &  0.591  &  0.831  &  0.691  &  0.701  \\
                    & World           & $\times$              &  0.814  &  0.723  &  0.628  &  0.673  &  0.754  \\ \bottomrule
\end{tabular}
\end{table}

\begin{table}[hbtp]
\small
\centering
\caption{Performance comparison of three global land cover maps against country-specific models for Togo based on the CropHarvest evaluation dataset \cite{kerner_rapid_2020, tseng2021cropharvest}. The multi-headed LSTM and TIML models metrics are taken from published results. The best results per metric are in \textbf{bold} and second best are \underline{underlined}.}
\begin{tabular}{@{}ccccc@{}}
\toprule
Map \textbf{/} Model  & Precision & Recall & F1-score & Accuracy \\ \midrule
ESA WorldCover 2020   & \textbf{0.875} & 0.528 & 0.659 & \underline{0.810}     \\
ESRI 2020             & 0.636 & 0.066 & 0.120 & 0.663     \\
Dynamic World 2020    & 0.609 & \textbf{0.764} & \underline{0.678} & 0.748     \\ \midrule
multi-headed LSTM \cite{kerner_rapid_2020}  & \underline{0.81}  & \underline{0.68}  & \textbf{0.74}  & \textbf{0.83}      \\ 
TIML \cite{tseng2022timl}    &  -    &  -    & 0.720  &  -         \\ \bottomrule
\end{tabular}
\label{tab:other_countries}
\end{table}

\end{document}